\pdfoutput=1

\documentclass[11pt]{article}

\usepackage[final]{coling}

\usepackage{times}
\usepackage{latexsym}

\usepackage[T1]{fontenc}

\usepackage{url}
\usepackage{soul}
\usepackage{microtype}
\usepackage{inconsolata}
\usepackage{multirow}
\usepackage[spacesep,definevectors]{easyvector}
\usepackage{amsmath}
\usepackage{mathtools}
\usepackage{xspace}
\usepackage{algorithm}
\usepackage{algpseudocode}
\usepackage{hyperref}

\usepackage{booktabs}
\usepackage{bbm}
\usepackage{tcolorbox}

\DeclareMathOperator*{\argmax}{arg\,max}

\newcommand{\OurDATA}{\textsc{MQA-Aeval}}
\newcommand{\OurMODEL}{\textsc{MQA-Keal}} 
\newcommand{\MQA}{\textsc{MQA-Ke}} 

\newcommand{\eat}[1]{}
\newcommand{\fixn}[1]{\footnote{\textcolor{red}{\textbf{FIX-Nawal!!!} #1}}}
\newcommand{\fixm}[1]{\footnote{\textcolor{red}{\textbf{FIX-MuTayabba!!!} #1}}}
\newcommand{\warn}[1]{\textcolor{red}{#1}}

\newcommand{\eg}{\emph{e.g.,}\xspace}

\newcommand{\jias}{Jias-13B}
\newcommand{\acegpt}{AceGPT-13B}
\newcommand{\gptthree}{\textsc{GPT-3.5-turbo-instruct}}
\newcommand{\gptfour}{GPT-4o-mini}

\renewenvironment{abstract}%
         {\centerline{\large\bf Abstract}%
          \begin{list}{}%
             {\setlength{\rightmargin}{0.6cm}%
              \setlength{\leftmargin}{0.6cm}}%
           \item[]\ignorespaces}%
         {\unskip\end{list}}

\title{\OurMODEL{}: Multi-hop Question Answering under Knowledge Editing for Arabic Language}

\author{Muhammad Asif Ali$^{*,1}$, Nawal Daftardar$^{1,2}$, Mutayyaba Waheed$^{3}$, \\
\textbf{Jianbin Qin\thanks{MA Ali and J Qin are co-corresponding authors.}$^{,4}$, and Di Wang$^{1}$}\\
  $^1$King Abdullah University of Science and Technology, KSA \\
  $^2$King AbdulAziz University, KSA; $^3$University of Science and Technology, China\\ $^4$Shenzhen University, KSA 
}

\begin{document}

\maketitle
\begin{abstract}
Large Language Models (LLMs) have demonstrated significant 
capabilities across numerous application domains. A key challenge 
is to keep these models updated with latest available information, 
which limits the true potential of these models. 
Although, there have been numerous attempts for LLMs' Knowledge Editing (KE), 
\textit{i.e.,} to update and/or edit the LLMs' prior knowledge and 
in turn test it \textit{via} Multi-hop Question Answering under 
KE (\MQA{}), yet these studies are primarily focused on English language.
In this paper, we extend~\MQA{} for Arabic language.
For this, we propose: 
\textsc{\textbf{\underline{M}}}ulti-hop 
\textsc{\textbf{\underline{Q}}}uestioning 
\textsc{\textbf{\underline{A}}}nswering under 
\textsc{\textbf{\underline{K}}}nowledge
\textsc{\textbf{\underline{e}}}diting for 
\textsc{\textbf{\underline{a}}}rabic 
\textsc{\textbf{\underline{l}}}anguage (\OurMODEL{}).
\OurMODEL{} stores knowledge edits as structured knowledge units in the 
external memory. In order to solve multi-hop question, it first uses 
task-decomposition to decompose the question into smaller sub-problems. 
Later, for each sub-problem it iteratively queries the external memory 
and/or target LLM in order to generate the final response.
In addition, we also contribute~\textsc{MQuAKE-ar} (Arabic translation 
of English benchmark~\textsc{MQuAKE}), as well as curate a new 
benchmark~\OurDATA{} for rigorous performance evaluation of \MQA{} 
for Arabic language. Experimentation evaluation reveals~\OurMODEL{} 
outperforms the baseline models by a significant margin.
\end{abstract}

\vspace{-1.7ex}
\section{Introduction}
\label{sec:intro}
\vspace{-1.7ex}
Large Language Models (LLMs) have demonstrated 
immense potential across a wide range of natural language 
applications~\cite{zhu2023multilingual, huang2023approximating, zhao2023survey}.
\eat{However, a key limitation of LLMs is their reliance on extensive 
training data requiring substantial training time and computational cost.}
A key challenge for these models is their limited
adaptability to recent events and/or new data. 
For instance, training data for Llama-2~\citep{2023_llama} only encompasses 
information about events till September 2022. This in turn restricts the true 
potential of these models to generate accurate responses about emerging 
events and questions beyond their training scope/timeline. 
\eat{Notably, it leads to hallucinations, a phenomenon 
when the LLMs try to generate plausible yet inaccurate 
responses for unknown facts~\cite{2023_hallucinate}.}

\begin{figure}[t]
    \centering
    \includegraphics[width=0.94\linewidth]{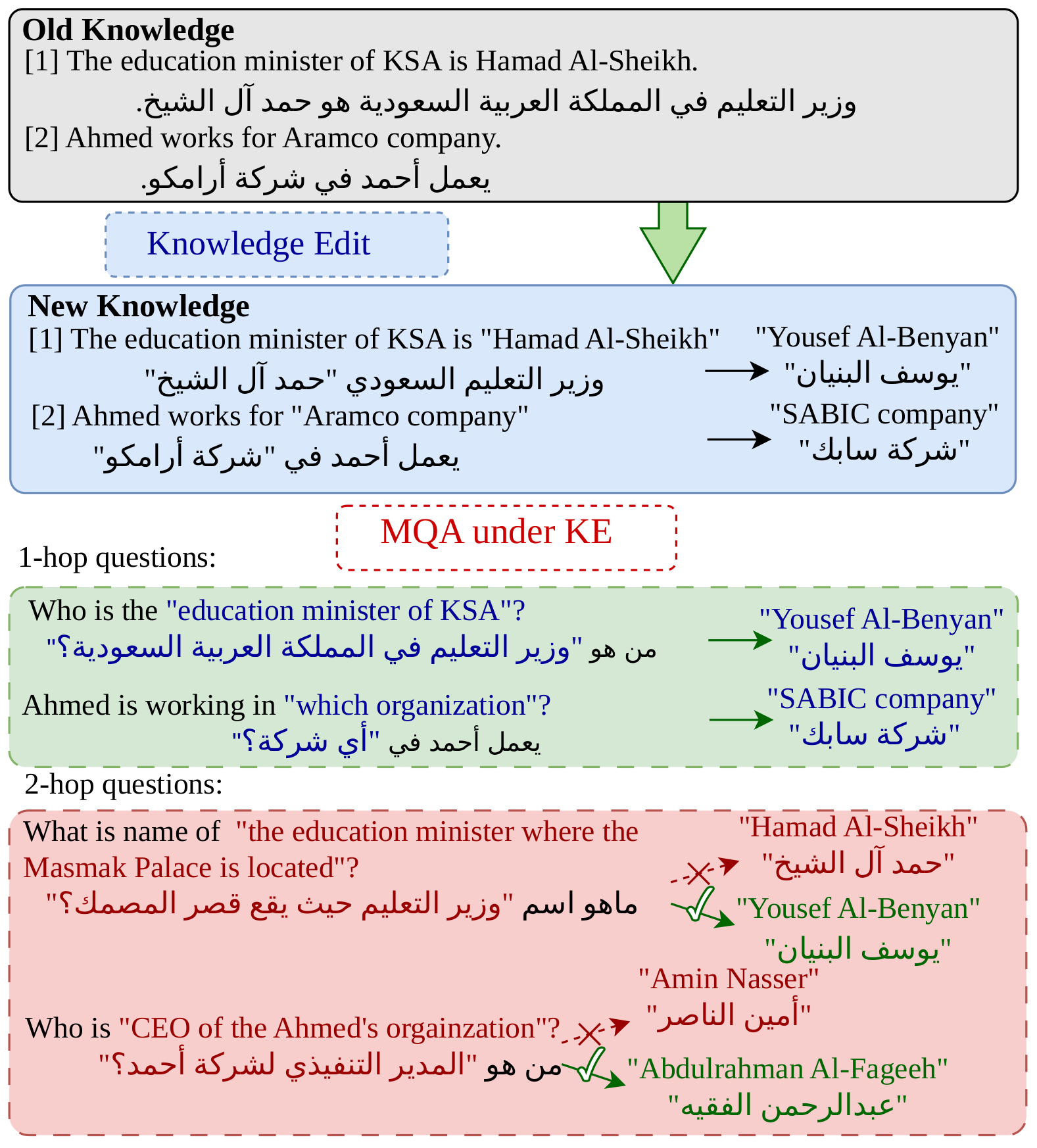}
    \vspace{-1.7ex}
    \caption{An example illustration of multi-hop question answering 
    under knowledge editing for Arabic and English language.}
    \vspace{-3.7ex}
    \label{fig:ex1}
\end{figure}

For this, numerous Knowledge Editing (KE) methods 
have been proposed that attempt to inject information
about new facts, while avoiding massive costs associated with model 
re-training~\cite{2021lora, Mitchell2021FastME, meng2022locating, meng2022mass}.
However, these methods do not provide a comprehensive solution for the KE problem.
For example, these models directly test the updated model for the edited 
knowledge without worrying about its impact on model's prior knowledge and/or 
facts explicitly correlated with the edits.
An example illustration in this regard is shown in Figure~\ref{fig:ex1}, 
which emphasizes that if we 
edit the information about the \textit{``Ahmed's workplace"}, 
corresponding knowledge/information about \textit{``Ahmed's 
Boss/CEO"} also needs to be updated, a phenomenon widely 
known in literature as \textit{``\textbf{ripple effect}''}.

To overcome these limitations, recently there have 
been numerous research attempts in order to design and develop robust 
KE methods and corresponding evaluation benchmarks that allow testing 
KE at multiple hops centered around the edit, also known as Multi-hop 
Question Answering under Knowledge editing (\MQA{}). 
Existing research on \MQA{} is primarily classified into 
parameter-based~\cite{2021lora,2023Ptuning} and memory-based 
variants~\cite{Mitchell2022MemoryBasedME, zhong2023mquake, gu2023pokemqa}, 
with memory-based methods outperforming the parameter-based methods.
We observe, that majority of the existing solutions for \MQA{} and their 
evaluation benchmarks are peculiarly tailored for English language.
While, recently LLMs have been extended to languages other than 
English, \eg AceGPT~\cite{2023_acegpt} for Arabic language;
Jias~\cite{2023_jais} a bilingual model supporting English 
and Arabic languages; and multi-lingual LLMs~\cite{2024multilingual}. 
There is a need to extend these methods to languages other than English. 

In this work we extend existing work on~\MQA{} to 
Arabic language. For this, we enumerate some of the key challenges as follows:
\underline{\textit{Firstly}}, existing best-performing memory-based 
solutions are inadequate, because these methods save edits as 
unstructured text embeddings in a shared memory, making it non-trivial to 
retrieve the correct edit for a given question. 
This situation gets worse, especially when the number of fact edits 
grow beyond a certain limit. 
\underline{\textit{Secondly}}, there is a need for an effective mechanism 
that can effectively correlate the edit with its most relevant 
part and/or sub-part in the question in order 
to augment the end-performance of the model. \underline{\textit{Thirdly}}, there is a need for 
appropriate evaluation benchmarks for a rigorous
evaluation of these systems for Arabic language.

Nevertheless, in this work we propose:
~\textsc{\textbf{\underline{M}}}ulti-hop 
\textsc{\textbf{\underline{Q}}}uestioning 
\textsc{\textbf{\underline{A}}}nswering under 
\textsc{\textbf{\underline{K}}}nowledge
\textsc{\textbf{\underline{e}}}diting for 
\textsc{\textbf{\underline{a}}}rabic 
\textsc{\textbf{\underline{l}}}anguage (\OurMODEL{}),
a novel approach, for \MQA{} in Arabic language.
\OurMODEL{} relies on following key components:
(a) \textit{``Structured Knowledge Retrieval"}, used to store the 
fact edits as a structured relational tuples in a shared memory.
(b) \textit{``Task-Decomposition"}, for decomposing the multi-hop 
questions into smaller sub-problems and/or knowledge units.
(c) \textit{``Iterative Traversal"} that  traverses over the sub-problem 
to generate a list of responses as candidate answers, as well as filtering 
the candidate answers by leverage logic rules in order to come up with 
the intermediate and/or the final response.

For evaluation, we use: 
(i)~\textsc{MQuAKE-ar}, an Arabic translation 
of an existing benchmark~\textsc{MQuAKE}.
(ii)~\OurDATA{}, a novel benchmark introduced in this 
work encompassing a wide range of single-hop and 
multi-hop questions primarily focused on Arabic Peninsula. 
Comprehensive experimental evaluation shows that~\OurMODEL{} 
outperforms the baseline models by a significant margin.  

We outline the key contributions of this work as follows:

\begin{enumerate}
\itemsep0em 
    \item We propose~\OurMODEL{}, a novel approach for 
    \textsc{MQA-Ke} for Arabic language that initially 
    decomposes multi-hop question into small sub-problems 
    to generate candidate answers, later leverages logic 
    rules to prune the 
    candidates to come up with the final response.
    
    \item We introduce two evaluation benchmarks, \textit{i.e.,} 
    \textsc{MQuAKE-ar} and \OurDATA{} for \MQA{} for Arabic language.
    
    \item We performed a comprehensive performance evaluation of~\OurMODEL{}, 
    showcasing that the proposed model outperforms the baseline models by a significant margin.
\end{enumerate}

\vspace{-1.7ex}
\section{Related Work}
\label{sec:RW}
\vspace{-1.7ex}

We classify the existing work on \MQA{} into: 
parameter based, and memory based methods.

The parameter based methods aim to fine-tune 
parameters of the large models in order to incorporate 
new knowledge and information.
Usually, fine-tuning is a highly time-consuming process 
and is also highly vulnerable to catastrophic forgetting, \textit{i.e.,} 
a phenomenon where model may forget and/or fail to retain its 
previous knowledge~\cite{Chen2020RecallAL}.
In order to avoid higher computational costs parameter-efficient 
variants were introduced. These models use an auxiliary set of 
parameters for fine-tuning, \textit{e.g.,} LoRA~\cite{2021lora}, 
Prompt Tuning~\cite{2023Ptuning}, QLoRA~\cite{2024_qlora}.

The memory based methods on the other hand store the edit 
information in an explicit memory, later use retrieval methods 
to retrieve the edit that is most relevant to the question.
Some examples include: 
SERAC by~\citet{Mitchell2022MemoryBasedME},
MeLLO by~\citet{zhong2023mquake},
PoKeMQA by~\citet{gu2023pokemqa}.

Usually, memory-based methods outperform 
the parameter-based methods. However, we 
observe, a key limitation of the memory-based 
methods is storing edits as embeddings learnt from 
unstructured text in a shared memory.
This makes it challenging to disambiguate among
different semantically relevant edits in the 
edit memory to retrieve the right fact edit.
This situation exacerbates especially, when the number 
of edits in the edit memory grow beyond a certain limit.
To overcome this~\OurMODEL{} stores edits as structured knowledge 
units, allowing relation-specific pruning \textit{etc.,} 
helpful to perform the end-task in a performance-augmented way.

\vspace{-1.7ex}
\section{Preliminaries}
\label{sec:prelimnaries}
\vspace{-1.7ex}
In this section, we introduce the mathematical 
notation and formulate our problem definition. 

\begin{figure*}[t]
    \centering
    \includegraphics[width=1.0\linewidth]{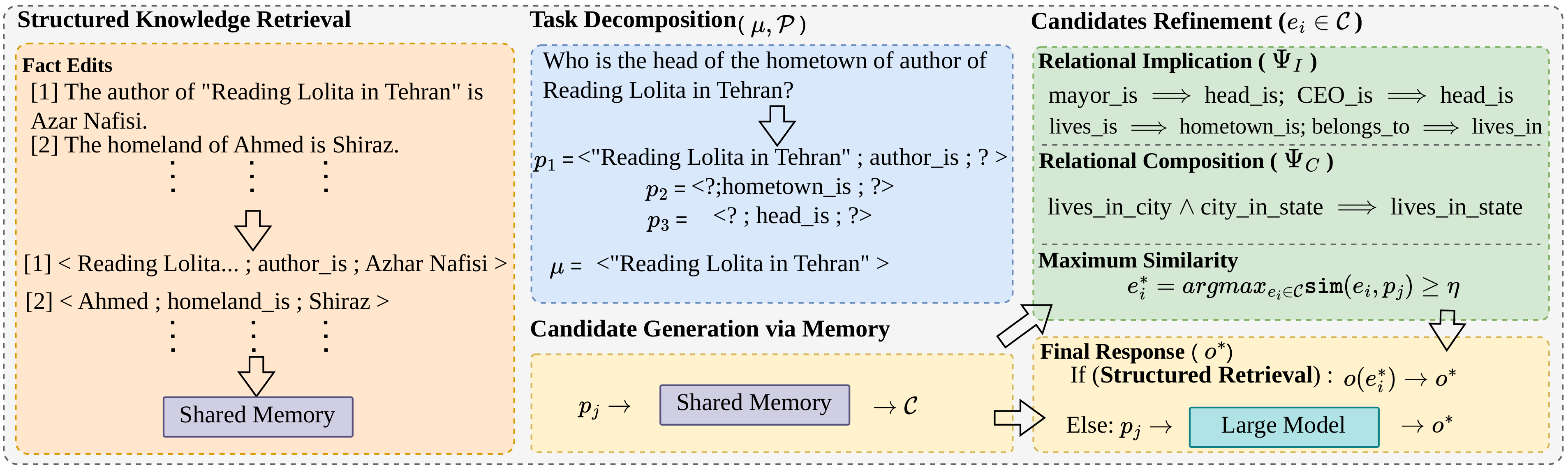}
    \vspace{-1.7ex}
    \caption{\textbf{Workflow of~\OurMODEL{}}. The left part of the Figure 
    shows how we store fact edits. The central part illustrates task decomposition and candidate generation from the memory. The right part of the Figure shows candidate refinement and final response generation.}
    \vspace{-2.7ex}
    \label{fig:framework}
\end{figure*}

\vspace{-0.7ex}
\subsection{Notation}
\vspace{-0.7ex}
For this work, we use knowledge graph triplets $(s,r,o)$ 
to represent the fact/knowledge, where $s$, $r$ 
and $o$ represent the subject, relation and object 
respectively. We use $e = (s,r,o \rightarrow o^{*})$
to represent an individual fact edit. It emphasizes 
that object of relation $r$ with subject $s$ is 
updated from $o$ to $o^{*}$. We use 
$\mathcal{E} = \{e_1,e_2,\cdots,e_n\}$ to represent the 
collection of edits.
We use $q \in \mathcal{Q}$ to represent multi-hop question. 
Answering $q$ requires multiple reasoning steps in 
order to compute the final response. 
We use 
$\mathcal{P} = \langle p_1,\cdots, p_n\rangle = \langle (s_1,r_1,o_1),\cdots, (s_n,r_n,o_n)\rangle$
to represent a chain of reasoning steps, with $o_n$ as the final 
response. 
For \MQA, if one of the fact undergoes an 
edit $e_i$, all subsequent facts in the knowledge 
chain needs to be updated. 
The resulting knowledge chain with updated path 
is represented as 
$\mathcal{P^{*}} = \langle (s_1,r_1,o_1),\cdots, 
(s_i,r_i,o_i \rightarrow o^{*}_i),\cdots, (s^{*}_n,r_n,o^{*}_n)\rangle$
with $o^{*}_n$ as the final answer. 
We $f(\cdot)$ to represent the target LLM.
\eat{We use T to represent the prompts and 
$f(\cdot)$ to represent the target LLM.}
We use $\psi_{i} \in \Psi_{I}$ and $\psi_{c} \in \Psi_{C}$ 
to represent the implication and compositional 
rules respectively.

\vspace{-0.7ex}
\subsection{Problem Definition}
\vspace{-0.7ex}
The task of knowledge editing is to update and/or modify the 
knowledge in the LLMs without fine-tuning the entire model.
Formally, given the LLM $f(\cdot)$, and a collection of 
fact edits $\mathcal{E} = \{e_1,e_2,\cdots,e_n\}$, we aim to 
augment the knowledge of $f(\cdot)$ using the facts edit 
information in $e_i \in \mathcal{E}$, such that it overrides 
the model's information about facts/knowledge 
correlated with $\mathcal{E}$, while keeping the other 
knowledge intact.
Later, use updated model to generate and/or deduce the 
final response $o^{*}_{n}$ for the multiple-hop question $q$.

\eat{update the 
Given the initial model parameters as $\theta$, the end-goal
of \OurMODEL{} is to update the model parameters to $\theta^{'}$
such that it overrides the information about knowledge $k$
while keeping the other knowledge intact.}


\vspace{-1ex}
\section{\OurMODEL{}}
\label{sec:proposed}
\vspace{-1ex}

\noindent{\bf Overview.}
In this paper, we propose
\textsc{\textbf{\underline{M}}}ulti-hop 
\textsc{\textbf{\underline{Q}}}uestioning 
\textsc{\textbf{\underline{A}}}nswering under 
\textsc{\textbf{\underline{K}}}nowledge
\textsc{\textbf{\underline{e}}}diting for 
\textsc{\textbf{\underline{a}}}rabic 
\textsc{\textbf{\underline{l}}}anguage (\OurMODEL{}), 
shown in Figure~\ref{fig:framework}.
\OurMODEL{} first uses 
\textit{``Structured knowledge Retrieval''}
to store the fact edits in a shared memory 
as structured relational triplets.
Later, it employs:
\textit{``Task Decomposition''}, to decompose 
the multi-hop question $q$ into sub-parts, and
\textit{``Iterative Traversal''} to traverse the
sub-parts to generate the response for $q$.
\eat{Note, the iterative traversal iterates over the
\textit{``Candidate Generation''}, and 
\textit{``Candidate Filtering''} phases in order 
to generate the final response.}
\eat{Further details about these model components 
is in the following sub-sections.}
Note, to the best of our knowledge this work 
is amongst the initial attempts for knowledge 
editing and in turn testing it via multi-hop 
question answering for the Arabic languages.

\vspace{-0.7ex}
\subsection{Structured knowledge Retrieval}
\vspace{-0.7ex}
In order to successfully answer the multi-hop 
questions the retriever must be able to understand 
and comprehend multiple information units requisite 
to accurately answer the question. In order to overcome 
the limitations posed by existing work that store edits as embeddings learnt 
over unstructured text, we store and retrieve fact edits as 
embeddings learnt over structured relational triplets. 
Underlying reason in this regard is the fact that usually 
the key information within an individual edit $e_i$ may be
primarily organized in form of relational triplets that 
allows relation-specific information filtering at later stages.

For example, the sentence: \textit{``The president of Iran 
is Masoud Pezeshkian"} may be organized as: 
<Iran; president\_is; Masoud Pezeshkian>.
At the same time, we can sub-divide the multi-hop question 
into smaller information units and/or sub-problems (details 
in Section~\ref{sec:task_decompose}), and accordingly
iteratively query the retriever with the sub-parts of the 
multi-hop question in order to generate the final response.

\begin{figure}[t]
    \centering
    \includegraphics[width=1.0\linewidth]{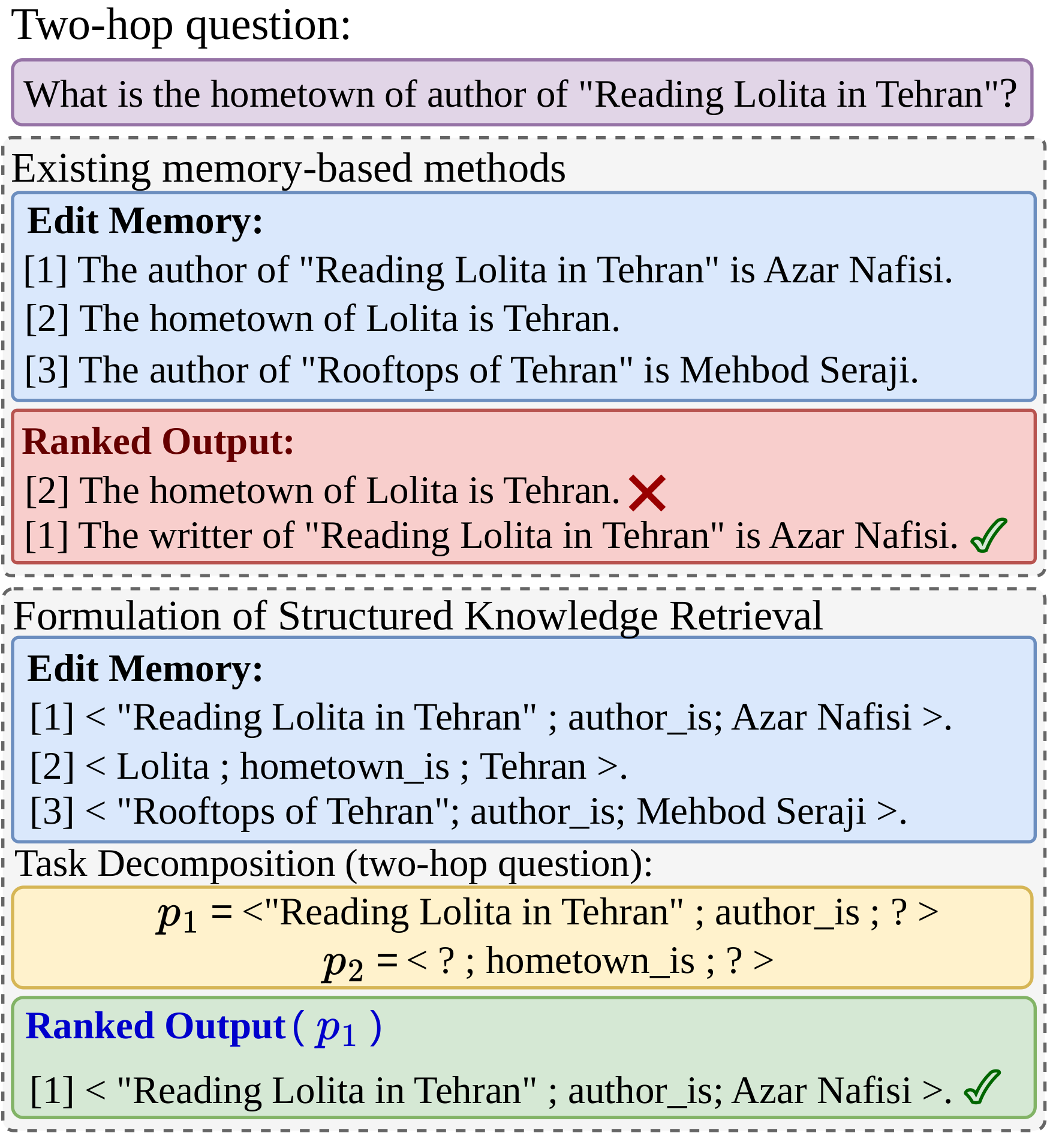}
    \vspace{-1.7ex}
    \caption{An example illustrating the limitation of existing 
    memory-based methods that store knowledge edits as unstructured 
    text, vs structured knowledge retrieval employed by~\OurMODEL{}.}
    \vspace{-2.7ex}
    \label{fig:memory}
\end{figure}

We also illustrate this phenomenon in Figure~\ref{fig:memory},
where the upper half of the Figure shows that for the 
two-hop question: 
\{\emph{``What is the hometown of author of Reading 
Lolita in Tehran''?}\}, the fact edit that is most
semantically similar to the question comes out to be:
\{\emph{``The hometown of Lolita is Tehran''}\}.
However, this retrieved fact does not guarantee 
whether we can use this information to successfully 
answer the question.
On the other hand, lower-half of the Figure shows
our formulation of structured knowledge retrieval,
that emphasizes if we store the edits as relational 
triplets, and accordingly query the model by 
decomposing the multi-hop question into sub-parts, 
we can accurately yield the edits that are relevant to 
each sub-part of $q$ iteratively.

Formally, for edits $e_i \in \mathcal{E}$, we decompose 
the edits as relational triplets $e_i = \; <s, r, o>$ 
and use a retrieval model, \textit{i.e.,} \texttt{Contriever}~\cite{2021_contriever}, to 
embed and save these edits as a retrieval index.
Later, during the model inference the index takes the query
as input and generates top-$k$ facts
most relevant to the input query.

\noindent{\bf \underline{Example}.} An example illustration 
of the structured memory retrieval module of~\OurMODEL{} is 
shown in the left half of Figure~\ref{fig:framework}, 
where the unstructured text 
\{\emph{``The homeland of Ahmed is Shiraz''}\}, is 
embed in the memory as 
\{\emph{$<$ Ahmed; homeland\_is; Shiraz $>$}\}.

\eat{The workflow of~\OurMODEL{} to generate the final
response is explained as follows:}

\vspace{-0.7ex}
\subsection{Task Decomposition}
\label{sec:task_decompose}
\vspace{-0.7ex}
Task decomposition module of~\OurMODEL{} aims to decompose 
the multi-hop question $q$ in to smaller sub-components in 
order to come up with a reasoning path and/or chain 
($\mathcal{P}$) that can be traversed iteratively 
in order to generate the final response. 

For this, we leverage the instruction following abilities 
of the LLMs to decompose the multi-hop question $q$ using an 
in-context learning prompt. Formally, we use the target 
model $f(\cdot)$ to decompose the $q$ as follows:

\vspace{-3.7ex}
\begin{equation}
\label{Eq:rel_extract}
     \mu, \mathcal{P}_\text{}= <p_1, p_2, \cdots p_n> = f(\text{T}_\text{relation}, q)
\end{equation}

\noindent here, $\text{T}_\text{relation}$ represents the in-context learning prompt used to decompose $q$, as outlined in Appendix~\ref{Appendix:Prompts}.
The output of the model is  
a chain of reasoning path indicative of 
the key components in $q$, \textit{i.e.,} $\mathcal{P}_\text{}$, and the starting point of the path traversal, \textit{i.e.,} $\mu$. 
Note, the start point $\mu$ is an entity and individual 
facts in $\mathcal{P}$ are organized in the form of relational 
triplets.

This formulation makes it convenient to iteratively 
traverse the $\mathcal{P}$ by either retrieving corresponding 
facts in the edit memory or querying the target 
LLM to come up with the final response.

\noindent{\bf \underline{Example}.} Continuing our previous 
example, the upper centre half of the Figure~\ref{fig:framework}
shows how~\OurMODEL{} decomposes a 3-hop question:
\{\emph{``Who is the head of the hometown of author of
Reading Lolita in Tehran''?}\} 
into $\mu$ = \emph{``Reading Lolita n Tehran''} and 
$\mathcal{P} = \{p_1, p_2, p_3\}$ with 
$p_1$ = <\text{\emph{``Reading Lolita in Tehran''}; \emph{author\_is}; ?}>, 
$p_2$ = <\text{? ; \emph{hometown\_is}; ?}>, and
$p_3$ = <\text{? ; \emph{head\_is}; ?}>. 

\vspace{-0.7ex}
\subsection{Iterative Traversal over $\mathcal{P}$}
\vspace{-0.7ex}
After decomposing the multi-hop question $q$ into 
multiple sub-problems, the iterative traversal part of 
~\OurMODEL{} iterates through the~$\mathcal{P}$
one step at a time with $\mu$ as the starting
point. During each step, it attempts to solve a smaller 
sub-problems \eat{($p_j \in \mathcal{P}$) }with results 
\eat{$(o_{i}^{*})$ }to be used as starting point for the next round. 
For this,~\OurMODEL{} repeatedly iterates through 
multiple rounds of response generation using shared 
memory and target LLM. 

Formally, for each sub-problem ($p_j \in \mathcal{P}$), 
the candidate generation modules looks for probable 
candidates $\{o_{i}^*\}$ for the answer. 
Given the fact, the end-goal of knowledge
editing is to update the priorly contained knowledge
of the LLMs. Thus, for response generation, we prioritize 
the responses retrieved from the from the edit memory, in case the 
information about a certain entity and/or facts has been 
updated (Section~\ref{sec:retrevial}). 
For cases, the model is not able to generate
substantial response from the edit memory, we resort
back to querying the target model to generate the final 
response (Section~\ref{sec:LLM_retrevial}).

\eat{
Note, unlike existing works that use logic rules to 
augment and/or extend the datasets, we leverage these rules
to prune candidate fact edits from the structured knowledge..!}

\subsubsection{Response from Structured Retrieval}
\label{sec:retrevial}
In contrast to the existing memory based methods~\citep{zhong2023mquake} 
that only consider top-most index semantically related 
to the input query, we retrieve top-$k$ edits as 
candidate answers, later refine these candidates via 
different pruning heuristics in order to generate the 
response $o_{i}^{*}$.
We argue, this formulation of retrieving 
response by selecting multiple candidates as 
probable answers helps in overcoming the limitations
posed by the memory-based systems and eventually
helps in significantly augmenting the end-performance 
of~\OurMODEL{}.

Formally, for a given sub-problem $(p_j \in \mathcal{P})$,
we look for the top-$k$ fact edits that are most relevant 
to the sub-problem, as follows:

\vspace{-3.7ex}
\begin{equation}
    \mathcal{C} = [e_1, \cdots, e_k] 
    = k\texttt{-}\!\!\!\argmax_{e_i \subset \mathcal{E}: |e_i| = k} 
\texttt{sim}(e_i, p_j)
\end{equation}
\vspace{-1.7ex}

\noindent where $k$-$\argmax$ returns the indices of the
top scored $k$ fact edits. $\texttt{sim()}$ is used to compute 
the embedding similarity of the embedding vectors\footnote{Note, we 
use dot product as a metric indicative of similarity among 
embedding vectors.}.
Finally, we consider $\mathcal{C}$ as the final set 
of candidates passed through the candidate filtering 
process, as detailed below.

\eat{For the case, we are able to get highly confident candidates 
from our structured retrieval memory, we directly move to the 
candidate refinement process.}

\begin{algorithm}[t]
    \caption{\scshape Candidate Filtering}
    \label{alg:alg1}
    \textbf{Input:} {$\eta:$ {thr}; $\mathcal{C} :$ {candidates}; 
    $p_j$; $\{p_1 \cdots p_n\} \in \mathcal{P}$}\\
    \textbf{Output:} {$o_{i}^{*}$}
    \begin{algorithmic}[1]
        \State {$o_{i}^{*} \gets \texttt{null}; \texttt{found} \gets \texttt{False}$}
        \For {$e_i \in \mathcal{C}$}\\
        \;\;\;\; \#1. Relational Implication ($\Psi_{I}$)
        \State {\#\# \textit{subject in alias check}}
        \If {$s(p_j) \in [\texttt{alias}(s(e_i))]$}
        \State {\#\# \textit{relational implication}}
        \If {$\psi_{i}\!:\!r(p_j) \rightarrow r(e_i)$} 
        \State {$o_{i}^{*} \gets o(e_i)$ \& $\texttt{found} \gets \texttt{True}$}
        \EndIf
        \EndIf \\
        \;\;\;\; \#2. Relational Composition ($\Psi_{C}$)
        \If {\texttt{found} == \texttt{False}}
        \If {$\psi_{c}\!:\!r(p_{1}) \land \! \cdots \! \land r(p_{j}) \rightarrow r(e_i)$}
        \State {$o_{i}^{*} \gets o(e_i)$ \& $\texttt{found} \gets \texttt{True}$}
        \EndIf
        \EndIf
        \EndFor \\
        \#3. Maximum Similarity
        \If {\texttt{found} == \texttt{False}}
        \State {$e_i^{*} = \argmax_{e_i \in \mathcal{C}_1} \texttt{sim}(e_i, p_j) \geq \eta$}
        \State {$o_{i}^{*} \gets o(e_i^{*})$}
        \EndIf
        \State {\Return $o_{i}^{*}$}
    \end{algorithmic}
\end{algorithm}

\noindent{\bf Candidate Filtering.}
The process-flow of candidate filtering is illustrated in
Algorithm~\ref{alg:alg1}.
It takes similarity threshold $\eta$,
list of candidate answers $\mathcal{C}$, 
the sub-problem $p_j \in \mathcal{P}$ as input,
and generates the final response $o_{i}^*$ as 
output for the sub-problem $p_j$.

For this, it initializes the variable $o_{i}^*$ to \texttt{null}, 
and uses a variable \{\texttt{found}\} initialized to \texttt{False}, 
used to keep track of the response generation.
Later, for each candidate (\textit{i.e.,} $e_i \in \mathcal{C}$),
the Algorithm~\ref{alg:alg1} iterates through three different 
stages, enumerated as follows:

\noindent{\underline{\textit{1. Relational Implication.}}}
This is outlined in lines (3-10) in Algorithm~\ref{alg:alg1}.
It considers the logical implication of relation pairs
in the sub-problem $(p_j)$ and candidate edit $(e_i)$ in 
order to capture semantically related relations.
We define relational implication as:

\noindent \textit{Definition:} \emph{For two relations,
$r_1$ implies $r_2$} (or $r_1 \implies r_2$) 
$\;\;\;\texttt{iff} \;\;\; \forall (s, o) \in r_1 \implies (s,o) \in r_2$ or $r_1 \subset r_2$. 

\noindent Some examples in this regard include: 
$\texttt{father\_of}$ $\implies$ $\texttt{parent\_of}$;  
$\texttt{lives\_in}$ $\implies$ $\texttt{belong\_to}$, 
\textit{etc.} For instance, if John is father of Tom, 
then John is also parent of Tom.

Formally, given two relations $r_1$ and $r_2$, we aim 
to compute whether $r_1 \implies r_2$.
For this, we first perform subject entity disambiguation by 
analyzing if the subject of the $p_j$, \textit{i.e.,} 
$s(p_j)$ matches with the alias names of subject in 
$e_i$, \textit{i.e.,} $[\texttt{alias}(s(e_i))]$, as 
outlined in lines (4-5).
Later, in lines (6-7) we look for implication of 
relation pairs, \textit{i.e.,} $r(p_j) \implies r(e_i)$
to assign the $o(e_i)$ as the answer $o_{i}^{*}$, as shown in line-8.
Details about the implication rule extraction are
explained in Appendix~\ref{Appendix:rule_implication}.

\noindent{\underline{\textit{2. Relational Composition.}}}
This is outlined in lines (11-17) in Algorithm~\ref{alg:alg1}.
It aims to compute the relational composition along the 
knowledge path $p_j \in \mathcal{P}$. 
For this, we use horn rule to compute the relational
composition, defined as follows.

\noindent \textit{Definition:} 
\emph{Horn rule is a special 
class of first-order logic rules that is composed of 
conjunctive predicates: 
$\mathbf{r_b} = \{r_{b_1}, \cdots r_{b_n}\}$ 
known as rule body or pre-condition, and a predicate ${r_h}$ 
as the rule head or consequence, represented as follows:}

$r_{b_1}(s,z_1) \wedge \cdots \wedge r_{b_n}(z_{n-1},o) \implies r_h(s,o)$

\noindent Some exemplar horn rules regard include: 
$\texttt{lives\_in\_city} \wedge \texttt{city\_in\_continent} 
\implies \texttt{lives\_in\_continent}$. This rules aims to 
look for the candidate fact edits $e_i \in \mathcal{C}$ that 
are strongly correlated with the fact chain $\{p_1, p_2 \cdots p_n\}$.

As mentioned in lines (13-14), if the relational predicates 
satisfy the precondition along with the relational part of 
the edit satisfying the con-sequence of the horn rule $\psi_{c}$, 
we use corresponding $o(e_i)$ as the answer $o_{i}^{*}$.
Further details about compositional rule extraction are explained 
in Appendix~\ref{Appendix:rule_composition}.

\noindent{\underline{\textit{3. Maximum Similarity.}}}
This step is outlined in lines (18-22) in Algorithm~\ref{alg:alg1}.
It aims to capture fact edit exhibiting higher embedding similarity 
with the sub-problem. 
For this, we select the fact edit $(e_i)$ exhibiting similarity 
with $p_j$, compared against a threshold $\eta$, \textit{i.e.,}
$\texttt{sim()} \geq \eta$ to assign the corresponding $o(e_i)$ 
as the response $o_{i}^{*}$.

\eat{For the cases, where we are not able to find fact edit, we assume 
there is no corresponding fact-edit relevant with the sub-problem $(p_j)$. 
For this, we solicit response from the target LLM, as explained below.}

\subsubsection{Response via target LLM}
\label{sec:LLM_retrevial}
For the cases,~\OurMODEL{} is not able to generate
substantial response from the structured knowledge
retrieval, we resort to the information contained 
in the target LLM in order to generate the response.
For this, we leverage the in-context learning 
abilities of the target LLM $f(\cdot)$, to generate 
a response for $p_j$, as follows:

\vspace{-3.1ex}
\begin{equation}
\label{Eq:llm_response}
     o_{i}^{*} = f(\text{T}_\text{query}, p_{j})
\end{equation}

\noindent where $p_j$ corresponds to the $j$-th 
sub-problem of the multi-hop question $q$,
and $\text{T}_\text{query}$ is the prompt used 
to query the target model (outlined in Appendix~\ref{Appendix:Prompts}).

The final response generated in this stage is 
used as an initial step for the subsequent steps to be 
traversed over the path $\mathcal{P}$, finally resulting 
in $o^{*}_n$ as the end response for the multi-hop question $q$.

\vspace{-1ex}
\section{Experimentation}
\label{sec:EXP}
\vspace{-1ex}

\subsection{Experimental Settings}

\paragraph{Datasets.}
In order to evaluate the performance of~\OurMODEL{},
we translated existing widely used data sets for MQA 
under KE,~\textit{i.e.,} \textsc{MQuAKE}~\citep{zhong2023mquake},
renamed as~\textsc{MQuAKE-ar} 
(encompassing~\textsc{MQuAKE-T-ar} and~\textsc{MQuAKE-CF-ar}). 
Note, these data sets were 
translated using automated tools and manually validated by local 
Arabic language experts. 
Apart from this, we also curated a new data set, namely~:
\underline{\textbf{M}}ulti hop 
\underline{\textbf{Q}}uestion
\underline{\textbf{A}}nswering 
under knowledge editing for 
\underline{\textbf{A}}rabic-region
\underline{\textsc{\textbf{eval}}}uation
(\OurDATA{}) that 
portray a more realistic setting for MQA under 
KE typically tailored for local Arab world.
Further details about the datasets, translation 
process and statistics of the data sets are 
provided in Appendix~\ref{Appendix_exp:dataset}.

\eat{We use a combination of self-curated 
data and existing benchmarks under single-hop and 
multi-hop settings.
Specifically, we curated ~\OurDATA{} that encompasses
information about recent events in Arabic Peninsula.
The evaluation benchmarks under single hop settings
include: Counterfact~\cite{wu2023evakellm} and 
KnowEdit~\cite{wang2023easyedit}.
1. Counterfact dataset\footnote{\url{https://rome.baulab.info/}},used in Eva-KELLM~\citep{wu2023evakellm}\\
2. KnowEdit \footnote{\url{https://huggingface.co/datasets/zjunlp/KnowEdit}} a combination of ZsRE, WikiBio, WikiDatacounterfact, Convsent and Sanitation. built by \citet{wang2023easyedit}
}

\noindent{\bf Baseline Models.}
We use existing best performing solutions for KE and \MQA{} 
as baselines. These include: 
(i) Fine-Tuning (FT)~\citep{Zhu2020ModifyingMI},
(ii) ROME~\citep{meng2022locating},
(iii) MEMIT~\citep{meng2022mass}, and
(iv) MeLLo~\citep{zhong2023mquake}.
Further details about the baseline models are 
provided in Appendix~\ref{Appendix:baseline}.

\noindent{\bf Evaluation Metrics.}
For performance evaluation, we use multi-hop accuracy (M-Acc)~\cite{zhong2023mquake}, 
and hop-wise accuracy (H-Acc)~\cite{gu2023pokemqa}.
Further details about the evaluation metrics and 
their mathematical formulation 
are provided in Appendix~\ref{Appendix:eval}.

\begin{table*}[t]
\centering
\resizebox{0.88\linewidth}{!}{
\begin{tabular}{cccccccccccccc}
\toprule[1.0pt]
\multirow{3}{*}{\textbf{Method}} & \multicolumn{6}{c}{\textbf{\textsc{MQuAKE-CF-ar}}} & \multicolumn{6}{c}{\textbf{\textsc{MQuAKE-T-ar}}}  \\ 
\cmidrule{2-13} 
& \multicolumn{2}{c}{1-edited}    & \multicolumn{2}{c}{100-edited}     & \multicolumn{2}{c}{3000-edited}  & \multicolumn{2}{c}{1-edited}  & \multicolumn{2}{c}{100-edited} & \multicolumn{2}{c}{1868-editted}  \\ 
\cmidrule{2-13} 
                        & M-Acc  &H-Acc    & M-Acc  &H-Acc   & M-Acc  &H-Acc   & M-Acc  &H-Acc   & M-Acc  &H-Acc & M-Acc  &H-Acc \\ 
\hline

\multicolumn{13}{c}{\scshape \jias} 
\\ \hline
FT                      & 11.30   & 2.10  & 1.40  & 0.05   & 0.01   & -     & 28.47 & 19.11 & 23.54 & 11.43 & 0.54   & 0.11    \\
ROME                    & 5.79    & 1.70  & 2.90  & 0.07   & 2.45   & 0.57  & 14.57 & 8.95  & 17.15 & 8.75  & 1.45   & 0.78    \\ 
MEMIT                   & 6.14    & 3.40  & 5.75  & 2.60   & 1.97   & 0.75  & 17.44 & 7.85  & 13.13 & 6.95  & 11.87  & 5.18    \\
MeLLo                   & \underline{15.35}   & \underline{7.58}  & \underline{14.50} & \underline{6.75}   & \underline{12.55}  & \underline{4.58}   & \underline{35.22} & \underline{24.38} & \underline{33.19} & \underline{24.15} & \underline{27.27}  & \underline{18.96}   \\
\OurMODEL{}             & \textbf{24.69}   & \textbf{13.15} & \textbf{22.05} & \textbf{14.47}  & \textbf{18.17}  & \textbf{14.32} & \textbf{47.21} & \textbf{35.90} & \textbf{45.27} & \textbf{36.30} & \textbf{42.89}  & \textbf{34.15}   \\
\hline
\multicolumn{13}{c}{\scshape \acegpt} \\ 
\hline
MeLLo                   & \underline{17.83} & \underline{9.78}  & \underline{15.32}  & \underline{4.97}  & \underline{13.25} & \underline{4.54}  & \underline{67.35} & \underline{45.44}  & \underline{57.12}   & \underline{41.21}  & \underline{39.17}   & \underline{33.21}    \\
\OurMODEL{}             & \textbf{28.13} & \textbf{18.11}  & \textbf{22.47}  & \textbf{12.51} & \textbf{19.75} & \textbf{8.95}  & \textbf{74.50} & \textbf{67.15}  & \textbf{69.10}   & \textbf{61.58}  & \textbf{63.84}   & \textbf{57.85}   \\ 
\hline
\multicolumn{13}{c}{\scshape \gptthree} \\ 
\hline
MeLLo                   & \underline{21.20}   & \underline{7.5}   & \underline{18.70}  & \underline{9.52} & \underline{15.07}  & \underline{5.69}   & \underline{72.37} & \underline{61.19}  & \underline{67.47}  & \underline{58.43}  & \underline{45.41}   & \underline{34.87}    \\
\OurMODEL{}             & \textbf{30.14}   & \textbf{24.59}   & \textbf{25.27}  & \textbf{21.95} & \textbf{22.50}  & \textbf{18.95}   & \textbf{79.52} & \textbf{73.46}  & \textbf{73.52}  & \textbf{66.23}  & \textbf{65.45}   & \textbf{57.33}    \\
\bottomrule[1.0pt]
\end{tabular} }
\vspace{-1.7ex}
\caption{\textbf{Experimental results for~\textsc{MQuAKE-ar}.} 
The result of~\OurMODEL{} (\textit{i.e.,} M-Acc and H-Acc) for 
different datasets and target LLMs compared against the baseline 
methods. For each target LLM, we boldface overall best scores 
with the second best underlined.}
\vspace{-2.7ex}
\label{tab:result1_mqaka_ar}
\end{table*}

\paragraph{Experimental Setup.}

For experimentation, we 
use~\gptthree\footnote{\url{https://platform.openai.com/}},
as well as existing Arabic-centric LLMs, \textit{i.e.,} 
\acegpt~\cite{2023_acegpt} and 
\jias~\cite{2023_jais}, as target LLMs. 
Further details about these LLMs are provided in
Appendix~\ref{Appendix:llm}.
Value of $\eta$ is set to 0.6 and 0.7 for~\textsc{MQuAKE-ar}
and~\OurDATA{} respectively.
Similar to MeLLO~\cite{zhong2023mquake}, 
for evaluation of~\OurMODEL{}, we used a batch of 
$k$ instances, \textit{i.e.,}
$k\in\{1,100, 3000\}$ for \textsc{MQuAKE-CF-ar},
$k\in\{1,100, 1868\}$ for \textsc{MQuAKE-T-ar},
and 
$k\in\{1,100, 229\}$ for \OurDATA{}.
In Section~\ref{sec:retrevial}, value of $k$=10 for top-$k$ 
candidate fact edits.
All experiments were repeated for five rounds, and 
average scores are reported.

\begin{table}[t]
\centering
\resizebox{1.02\columnwidth}{!}{
\begin{tabular}{cccccccc}
\toprule[1.0pt]
\multirow{3}{*}{\textbf{Method}} & \multicolumn{6}{c}{\textbf{\textsc{\OurDATA{}}}}  
\\ \cmidrule{2-7} 
& \multicolumn{2}{c}{1-edited}    & \multicolumn{2}{c}{100-edited}     & \multicolumn{2}{c}{229-edited}         
\\ \cmidrule{2-7} 
& M-Acc  &H-Acc    & M-Acc  &H-Acc   & M-Acc  &H-Acc    
\\ \hline
\multicolumn{7}{c}{\scshape \jias} 
\\ \hline
FT             & 5.74  & 1.15  & 2.24  & 0.95 & 1.95  & 0.05 \\
ROME           & 2.34  & 0.07  & 3.45  & 0.98 & 3.95  & 0.85 \\
MEMIT          & 3.45  & 0.45  & 4.85   & 0.55  & 4.54   & 1.10 \\
MeLLo          & \underline{24.42} & \underline{13.45}  & \underline{23.25}  & \underline{17.58}  & \underline{22.65}  & \underline{17.07} \\
\OurMODEL{}    & \textbf{37.64} & \textbf{29.45} & \textbf{35.32}  & \textbf{28.53} & \textbf{34.38}  & \textbf{24.45} \\
\hline
\multicolumn{7}{c}{\scshape \acegpt} 
\\ \hline
MeLLo          & \underline{27.50}  & \underline{20.21}  & \underline{25.13}  & \underline{17.59} & \underline{24.17}  & \underline{18.01} \\
\OurMODEL{}    & \textbf{41.85}  & \textbf{31.59}  & \textbf{39.51}  & \textbf{27.31} & \textbf{38.32}  & \textbf{30.45} \\ 
\hline
\multicolumn{7}{c}{\scshape \gptthree} 
\\ \hline
MeLLo          & \underline{35.57}  & \underline{23.41}  & \underline{33.51}  & \underline{28.87} & \underline{32.64}  & \underline{26.45} \\
\OurMODEL{}    & \textbf{44.17}  & \textbf{39.45}  & \textbf{42.91}  & \textbf{37.41} & \textbf{41.65}  & \textbf{31.81} \\
\bottomrule[1.0pt]
\end{tabular} }
\vspace{-1.7ex}
\caption{\textbf{Experimental results for~\OurDATA{}.} 
We report the scores of~\OurMODEL{} (\textit{i.e.,} M-Acc and H-Acc), 
compared against baseline models. We boldface overall best scores 
with second best underlined.}
\vspace{-2.7ex}
\label{tab:result2_mqake_ar}
\end{table}

\vspace{-0.7ex}
\subsection{Main Results}
\vspace{-0.7ex}
The results of~\OurMODEL{} for~\textsc{MQuAKE-ar} and different target LLMs 
are shown in Table~\ref{tab:result1_mqaka_ar}. Here, we report M-Acc and 
H-Acc scores of~\OurMODEL{} compared against the baseline models. 
Analysing the results for the baseline models (\textit{i.e.,} FT, ROME, MEMIT), 
we observe that widely used knowledge editing methods perform poorly on~\MQA{}, 
which showcases the true knowledge augmentation potential of these
models for LLMs.

Overall results in Table~\ref{tab:result1_mqaka_ar} show that~\OurMODEL{} 
consistently outperforms the baseline models by a significant margin across both metrics.
For instance, for~\textsc{MQuAKE-CF-ar} using \gptthree{} as the target
LLM~\OurMODEL{} improves the M-Acc scores by \{42.1\%, 35.1\% and 49.3\%\}
respectively for \{1, 100 and 3000\} edited cases.
Likewise, for~\textsc{MQuAKE-T-ar}, and \gptthree{} as target LLM, the 
improvement in performance is \{9.8\%, 8.9\% and 44.1\%\} respectively
for \{1, 100 and 1868\}-edited cases.
The results of~\OurMODEL{} with \jias{} and \acegpt{} as target LLM exhibit 
similar behaviors with our proposed model outperforming the best performing 
baseline models.

Analyzing the results of~\OurMODEL{} for newly proposed data 
(\textit{i.e.,}~\OurDATA{}) in Table~\ref{tab:result2_mqake_ar} 
shows a similar behaviour with~\OurMODEL{} yielding better performance 
than the baseline models. 
However, we observe the model performance on~\OurDATA{} is relatively
lower compared to that of~\textsc{MQuAKE-T-ar}. 
We enumerate some of the probable justifications in this regard as follows:
(i) the instances in~\OurDATA{} are localized for local Arabic region 
which may not be adequately covered in LLMs' training corpora, as most of 
the existing LLMs are trained using knowledge directly acquired and/or translated 
from western regions~\cite{2023_culture_bias};
(ii) majority of the information in~\OurDATA{} (and corresponding reasoning 
path~$\mathcal{P}^{*}$) is about recent events, beyond the cut-off of LLMs 
training data; and
(iii) it is not easy to get aliases for entities for Arabic language, 
which limits the entity matching abilities of~\OurMODEL{} 
(line-5 in Algorithm~\ref{alg:alg1}).

We observe, amongst all data sets,~\OurMODEL{} yields best scores 
for~\textsc{MQuAKE-T-ar} followed by~\OurDATA{} and~\textsc{MQuAKE-CF-ar}.
This behavior is also consistent with the baseline models.
\eat{A possible justification in this regard is the fact that~\textsc{MQuAKE-CF-ar}
offers a rigorous and comprehensive evaluation platform also allowing 
multiple numbers of edits for evaluation.}
Analysing the results for different target LLMs, we observe~\gptthree{} yields 
best performance overall followed by~\acegpt{} and~\jias{}.
A possible reason in this regard is the fact that~\acegpt{} being 
an instruction-tuned variant of {Llama-2}~\cite{llama_2} inherits
better task-decomposition abilities compared to that of~\jias{}.
We also observe that with higher number of fact edits, the performance 
of the model decreases. 
This is attributable to multiple different factors as analyzed in 
Section~\ref{sec:error_cases}.
However, this effect is more pronounced for the baseline models compared 
to that of~\OurMODEL{}.

\vspace{-1.7ex}
\section{Analyses}
\vspace{-1.7ex}
In this section, we perform a detailed analyses of~\OurMODEL{} 
under different settings. 
This includes: 
(i) Ablation Analyses
(ii) Error Analyses.
Note, some additional experimental analyses are also 
reported in Appendix~\ref{Appendix_exp:experiments}.

\begin{table}[t]
\centering
\resizebox{1.02\columnwidth}{!}{
\begin{tabular}{cccccccc}
\toprule[1.0pt]
\multirow{3}{*}{\textbf{Method}} & \multicolumn{6}{c}{\textbf{\textsc{\textsc{MQuAKE-T-ar}}}}  
\\ \cmidrule{2-7} 
& \multicolumn{2}{c}{1-edited}    & \multicolumn{2}{c}{100-edited}     & \multicolumn{2}{c}{1868-edited}         
\\ \cmidrule{2-7} 
& M-Acc  &H-Acc    & M-Acc  &H-Acc   & M-Acc  &H-Acc    \\ 
\hline
\multicolumn{7}{c}{\scshape \gptthree} \\ 
\hline
\OurMODEL{} (--I)    & 71.31  & 67.45  & 65.38  & 61.57 & 59.55  & 52.10 \\
\OurMODEL{} (--C)    & 77.44  & 71.56  & 72.87  & 64.33 & 62.58  & 55.19 \\
\OurMODEL{} (--IC)   & 68.45  & 64.31  & 61.45  & 57.24 & 55.01  & 50.48 \\
\OurMODEL{}          & 79.52  & 73.46  & 73.52  & 66.23 & 65.45  & 57.33 \\

\bottomrule[1.0pt]
\end{tabular} }
\vspace{-1.7ex}
\caption{\textbf{Ablation analysis for~\textsc{MQuAKE-T-ar}} 
under varying conditions and~\gptthree{} as target LLM.}
\vspace{-2.7ex}
\label{tab:ablation}
\end{table}

\vspace{-0.7ex}
\subsection{Ablation Analyses}
\vspace{-0.7ex}
Ablation analysis aims to analyze the performance attributable to 
individual model components. For this, we report the performance 
of~\OurMODEL{} for 
(i) --I (\textit{w/o} implication rules),
(ii) --C (\textit{w/o} compositional rules),
(iii) --IC (\textit{w/o} both implication and compositional rules).

Corresponding results of~\OurMODEL{} with~\gptthree{}
as target LLM and~\textsc{MQuAKE-T-ar} data set are shown in 
Table~\ref{tab:ablation}. These results show that
omission of implication rules have a more pronounced impact on the model 
performance compared to that of the compositional rules.
For instance, compared with the complete model,
the variant~\OurMODEL{}(--I) exhibits \{10.3, 11.1, 9.0\}\% 
reduction in performance;
whereas,~\OurMODEL{}(--C) results in \{2.6, 0.8, 4.3\}\% 
decline in performance for \{1, 100, and 1868\}-edited cases respectively. 
This is understandable owing to the fact that overall implication
rules have a broader coverage and are more likely to be satisfied
compared to that of the compositional rules.
Also, jointly omitting the implication and compositional rules
exhibits a compounding effect, as evident in Table~\ref{tab:ablation} 
with~\OurMODEL{}(--IC) showing an accumulated decrease of 
\{13.9, 16.4, 15.9\}\% in M-Acc scores for \{1, 100, and 1868\}-edited cases.

Also, comparing the results of~\OurMODEL{}(--IC) with last row in
Table~\ref{tab:result1_mqaka_ar}, we observe that even if we omit 
the candidate filtering, the end-results of~\OurMODEL{} are still 
better than MeLLo, \textit{i.e.,} 
M-Acc = 55.01 for \OurMODEL{}(--IC) compared with M-Acc = 45.41 
for MeLLo under 1868-edited cases.
To summarize, these results show that the \emph{``structured knowledge reterieval''}
employed by~\OurMODEL{} followed by candidate filtering offer 
a robust setting helpful in performing the end~\MQA{} task in a 
performance enhanced way.

\subsection{Error Cases.}
\label{sec:error_cases}
We analyzed a random sample of 50 error cases of~\OurMODEL{}
in order to understand and comprehend the limitations of the model.
We categorize these error cases into following different categories: 
(a) errors by target LLMs,
(b) errors by structured retrieval, 
(c) errors in task decomposition by LLM, and
(d) miscellaneous errors.

For the variant of~\OurMODEL{} with~\OurDATA{} and 
\acegpt{} as the target model, we observe that almost 11\% of the 
errors were caused by the erroneous response generated by the 
target LLM (most probably because model is ignorant about some 
recent events beyond its training scope), 
25\% of the errors are caused by the structured fact 
retrieval, and 17\% errors were caused by in-appropriate 
task-decomposition by LLM.
Rest of the errors were categorized to miscellaneous errors.
This analysis shows that although our formulation of task 
decomposition along with structured knowledge retrieval employed 
by~\OurMODEL{} were able to significantly improve the performance 
compared to the unstructured text embeddings, yet it did not 
completely eradicate the issue.

\vspace{-1.7ex}
\section{Conclusions}
\vspace{-1.7ex}
In this work, we proposed~\OurMODEL{}, a novel approach 
for Knowledge Editing and in turn test the edited knowledge
via multi-hop question answering for Arabic language.
Apart from that, we also contributed~\textsc{MQuAKE-ar}
an Arabic translated and humanly-validated version of
existing \textsc{MQuAKE}~\cite{zhong2023mquake} data set; 
and~\OurDATA{} as new \MQA{} data set targeting 
information primarily on Arabian Peninsula.

\vspace{-1ex}
\section*{Limitations}
\vspace{-1ex}
Some of the core limitations of the proposed approach are outlined as follows:

\begin{itemize}
\itemsep0em
    \item \OurMODEL{} uses an iterative approach for 
    generating response for multi-hop questions. Errors 
    in the intermediate path may propagate and impact the final 
    answer. Currently, the is no effective mechanism for 
    recovery from errors in the intermediate path.
    
    \item Our work assumes, for the cases where there 
    is no fact edit directly and/or indirectly correlated 
    with a particular entity,~\OurMODEL{} entirely relies 
    on the target output. For these cases, the end-result 
    will be incorrect if the target model yields incorrect 
    results.
    
    \item For performance comparison,~\OurMODEL{} uses 
    corresponding target LLM to decompose the multi-hop 
    question into smaller sub-problems. For cases with 
    target LLM exhibiting relatively inferior knowledge 
    decomposition abilities, the end-result of~\OurMODEL{} 
    is severely impacted.
\end{itemize}


\section*{Ethics Statement}
This work fully complies with the \href{https://www.aclweb.org/portal/content/acl-code-ethics}{ACL Ethics Policy}. 
To the best of our knowledge, this study does not raise any 
ethical concerns and/or issues.

\bibliography{custom}

\clearpage
\appendix

\section{Background}
\label{Appendix:background}

\subsection{Knowledge Representation}
We use graph triplets $(s,r,o)$ to represent the fact/knowledge,
where $s$, $r$ and $o$ represent the subject, relation and object 
respectively. This is commonly used to represent the facts in 
the Knowledge graphs.

\subsection{Knowledge Editing (KE)}
\label{Appendix:KE}
We use $e = (s,r,o \rightarrow o^{*})$ to represent an individual 
knowledge update. It emphasize that the object of subject $s$
under relation $r$ is updated from $o$ to $o^{*}$. 
A collection of knowledge edits is represented by 
$\mathcal{E} = \{e_1, e_2, \cdots, e_n\}$.

\subsection{Multi-hop Question Answering under KE}
\label{Appendix:MQA}
A multi-hop question $q$ requires multiple reasoning steps 
in order to come up with the final answer/response. Generally, 
these reasoning steps formulate a chain of facts
$\mathcal{C} = \langle (s_1,r_1,o_1),\cdots, r_n(s_n,r_n,o_n)\rangle$.
The consecutive facts in $\mathcal{C}$ are chained together, i.e., 
$o_i$ from the proceeding step is $s_{i+1}$ for the subsequent fact, 
with $o_n$ as the final outcome.
For multi-hop question answering under KE, if one of the fact
$(s_i,r_i,o_i)$ in $\mathcal{C}$ undergoes an edit 
$(e_i \in \mathcal{E})$, the resulting chain for the subsequent 
facts need to be updated. The updated chain becomes:
$\mathcal{C} = \langle (s_1,r_1,o_1),\cdots, (s_i,r_i,o_i 
\rightarrow o_i^*),\cdots, (s_n^*,r_n, o_n \rightarrow o_n^*)\rangle$,
with $o^{*}_n$ as the final outcome.
The end-goal of multi-hop question answering is to come 
up with the answer for $q$ based on edits in $\mathcal{E}$.

Multi-hop question answering under knowledge editing
is a key challenge for LLMs. Some illustrative examples 
are shown below:

\eat{\section{Datasets Details}
\label{Appendix:data}
\subsection{\OurDATA{}}
\subsection{\textsc{MQuAKE-ar}}
It consists of two parts, \textit{i.e.,}
\subsection{\textsc{AToKE-ar}}
\cite{Yin2023HistoryMT}
\fixn{Add details tell SE, ME and other parts of the data}
\fixm{Add details about statistics of the data}}

\section{Rule Extraction}
\label{Appendix:rule_ext}
This work uses implication and compositional rules 
for filtering candidate response from structured memory 
for multi-hop questions. Details about the rule discovery
process are explained as below:

\subsection{Implication Rules}
\label{Appendix:rule_implication}
For implication rule mining, we used translated the existing set 
of implication rules provided by~\citep{2015_implication2, 2019_implication1}.
The translated rules were manually validated by local domain experts.

\subsection{Compositional Rules}
\label{Appendix:rule_composition}
For compositional rule mining, we use RNNLogic~\cite{2020_rnnlogic}
as our mining tool, to extract a set of compositional 
logic rules $(\Psi_{C})$ over Arabic Wikipedia\footnote{\url{https://dumps.wikimedia.org/arwiki/}}. 
After rule mining, we use rule's support threshold $(A_{\Psi_{C}})$ as a 
criterion for rule selection.

\begin{table}[t]
\centering
\resizebox{\columnwidth}{!}{%
\begin{tabular}{llllll}
\hline
Data                                   & \#Edits & 2-hop & 3-hop & 4-hop & Total \\
\hline
                                       & 1       & 2454  & 855   & 446   & 3755  \\
                                       & 2       & 2425  & 853   & 467   & 3745  \\
\textsc{MQuAKE-CF-ar} & 3       &       & 827   & 455   & 1282  \\
                                       & 4       &       &       & 436   & 436   \\
                                       & All     & 4879  & 2535  & 1804  & 9218  \\
\hline
\textsc{MquAKE-T-ar}  & 1 (All) & 1421  & 445   & 2     & 1868  \\
\hline
\OurDATA{}            & 1 (All) & 209     & 10     & 10     & 229 \\
\hline
\end{tabular}%
}
\caption{Statistics of data sets.}
\label{tab:dataset}
\vspace{-2.7ex}
\end{table}

\section{Additional Experimental Settings}
\label{Appendix_exp:experiments}

\subsection{Dataset}
\label{Appendix_exp:dataset}
For experimental evaluation, we curate an Arabic dataset
named:~\OurDATA{} encompassing information about recent 
events in the Arabian Peninsula.
Apart from that, we also use existing knowledge editing
benchmarks under single-hop and multi-hop settings. 
These data sets were translated to the Arabic 
language followed by validation from native speakers.
The statistics of the data set are given in 
Table~\ref{tab:dataset}, their details are as follows:

\noindent{\bf (i)~\textsc{MQuAKE-ar}.} 
\textsc{MQuAKE-ar} is the Arabic translated data of the o
riginal \textsc{MQuAKE} data by~\citep{zhong2023mquake}. 
We translate both components of~\textsc{MQuAKE},
\textit{i.e.,} \textsc{MQuAKE-CF} and \textsc{MQuAKE-t}.
\textsc{MQuAKE-CF-ar} include 3,000 k-hop questions 
($k\in{2,3,4}$) based on counter-factual editing. 
\textsc{MQuAKE-t-ar} is based on real-world knowledge 
changes to construct edit, but not given time scope.
The statistics of data set is given in Table~\ref{tab:dataset}.
For data translation, we use a semi-automated pipeline 
similar to the one used by~\cite{2024_bimedix}, 
\textit{i.e.,} using a two step 
process: 
(i) iterative translation and scoring using LLMs 
(\textit{e.g.,} ChatGPT),
(ii) manual refinement of low scored samples as well
as random samples from high-scoring samples.

\noindent{\bf (ii)~\OurDATA{}.}
Given that~\OurMODEL{} is focused on knowledge
editing and corresponding multi-hop question answering 
for Arabic language. 
Thus, in order to rigorously test the performance of
~\OurMODEL{}, we curated a new data set, namely:
\underline{\textbf{M}}ulti hop 
\underline{\textbf{Q}}uestion
\underline{\textbf{A}}nswering 
under knowledge editing for 
\underline{\textbf{A}}rabic-region
\underline{\textsc{\textbf{eval}}}uation
(\OurDATA{})
The statistics of data set is given in Table~\ref{tab:dataset}.
An instance of~\OurDATA{} is illustrated in the 
following example:

\noindent{\bf Example.}
An example instance of our newly generated data~\OurDATA{}
is shown in Figure~\ref{fig:ourdata_example}, given below.
\begin{figure}[h]
    \centering
    \includegraphics[width=1.0\columnwidth]{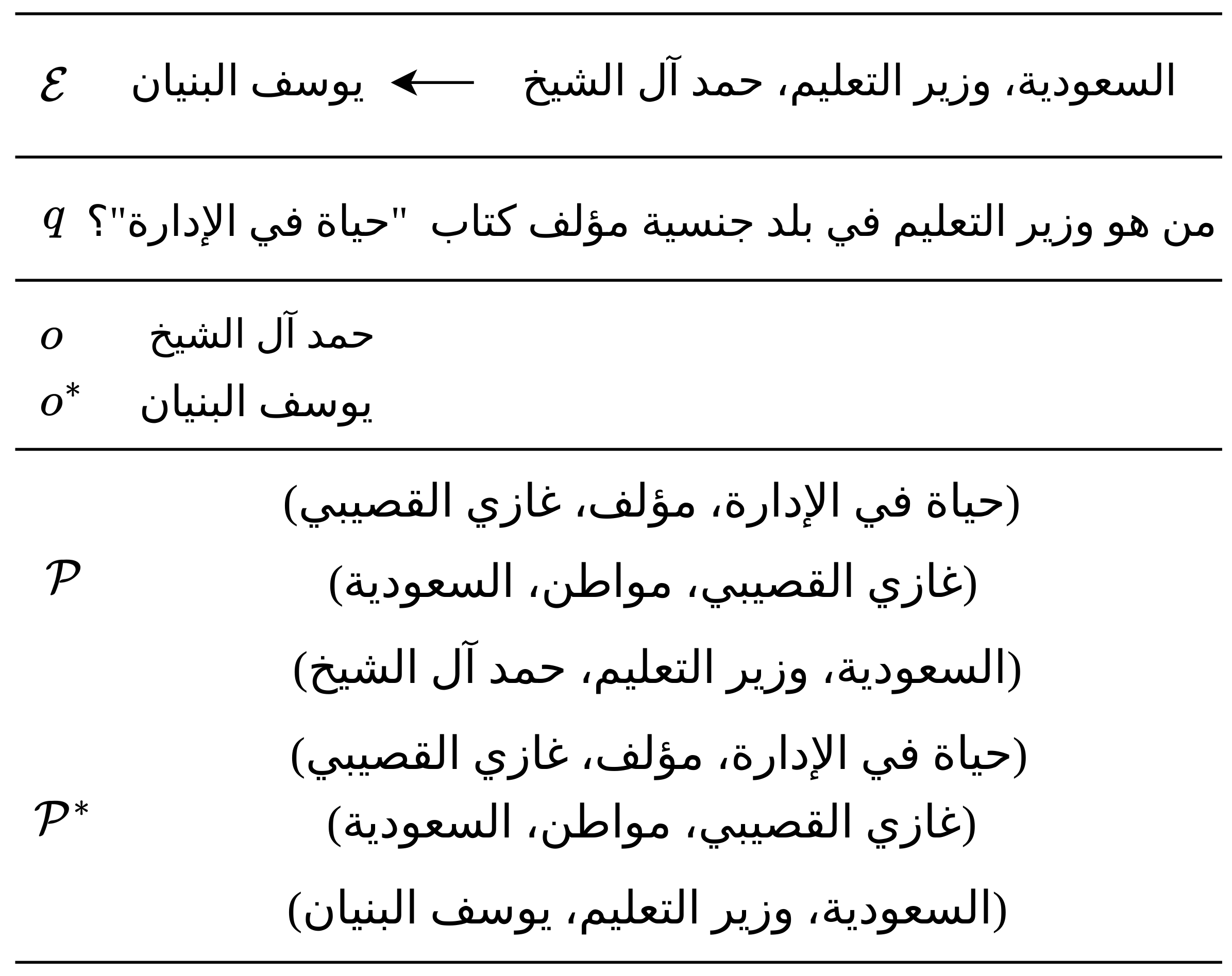}
    \vspace{-3.7ex}
    \caption{An example illustration of~\OurDATA{}.}
    \vspace{-2.7ex}
    \label{fig:ourdata_example}
\end{figure}

\subsection{Baseline Models}
\label{Appendix:baseline}
The details about the baseline models are provided as
follows:

\paragraph{(i) Fine-tuning (FT).} 
It uses updated/edited knowledge to fine-tune the
model parameters through gradient descent~\citep{Zhu2020ModifyingMI}.
\eat{utilize edited knowledge to fine-tune the parameters of 
model through gradient decrease.}

\eat{\paragraph{(b) MEND.} MEND employs a hyper-network to generate 
weight adjustments by decomposing the gradient obtained through 
conventional fine-tuning into low-ranks~\citep{Mitchell2021FastME}.}

\paragraph{(ii) ROME.}
ROME is based on the assumption that knowledge 
is stored in the feed-forward layers of transformer network. And, we can incorporate new knowledge in 
the model by simply locating and updating the 
parameters of these layers~\citep{meng2022locating}.

\paragraph{(iii) MEMIT.} MEMIT extends ROME by allowing a multi-edit scenario by editing multiple layers of model~\citep{meng2022mass}. 
\eat{enable high-accuracy on multiple edit scenario through extending ROME to edit feed-forward network on multiple.}

\eat{
\paragraph{() SERAC.} SERAC constructs a 
counterfactual model by preserving the original 
model and incorporating a classifier to 
decide whether to utilize the counterfactual 
model for responding to queries~\citep{Mitchell2022MemoryBasedME}.}

\paragraph{(iv) MeLLo.} MeLLo is a memory-based 
system to store the edited facts in an explicit 
memory, and prompts LLM to generate final 
response consistent with the edited facts
~\citep{zhong2023mquake}.

\subsection{Large Models}
\label{Appendix:llm}
We use existing Arabic centric language models

\paragraph{(a) \gptthree{}.}
\gptthree{} is released by Open-AI\footnote{\url{https://platform.openai.com/}},
in November 2023. This model uses a context 
window of 16,385 tokens and is trained till 
on a training data with a cut-off 
date of September, 2021.

\eat{\paragraph{(b) \gptfour{}.}
This one of the most advanced model released
by Open-AI in May 2024. This model is 
trained on the training data with a cut-off 
date of October, 2023.}

\paragraph{(b) \jias.} \jias{} is a state-of-the-art 
Arabic-centric generative language model trained 
on a mixture of Arabic, English and programming 
languages text~\citep{2023_jais}. 
We use foundation model with 13 billion parameters.

\paragraph{(c) \acegpt{}.} \acegpt{} is an attempt 
to incrementally pre-train existing LLMs using Arabic 
data to incorporate Arabic grammar, culture and 
values~\citep{2023_acegpt}. We use 13 billion 
variant for foundation model.

\subsection{Evaluation Metrics}
\label{Appendix:eval}

Details about the evaluation metrics and their mathematical formulation are provided as follows: 

\eat{\noindent{\bf (a) Edit-wise Accuracy (E-Acc)}. E-Acc aims to compute 
the edit-wise success rate, \textit{i.e.,} how many edited facts may 
be recalled by the edited model.
Formally, given the edit $e = (s,r, o \rightarrow o^*)$, we compute 
the edit-wise accuracy as $\mathbbm{1}[f^{*}(s,r) \text{=} o^{*}]$.
For E-Acc, we take average values of all edits in the data sets.

\noindent{\bf (b) Instance-wise Accuracy (I-Acc)}. I-Acc aims to compute 
the number of instances where all the associated facts may be recalled 
by the language model (original and/or edited model).
Formally, given a data instance 
$d = (\mathcal{E}, q, o, o^{*}, \mathcal{P}, \mathcal{P}^{*})$,
we compute instance wise accuracy for the base model $f()$ as:

\begin{equation}
\mathbbm{1}\left[\bigwedge_{(s,r,o) \in \mathcal{C}} [f(s,r) = o]\right].
\end{equation}

\noindent Likewise, instance-wise accuracy for the edited model $f^*(\cdot)$ is 
computed as:

\begin{equation}
\mathbbm{1}\left[\bigwedge_{(s,r,o^{*}) \in \mathcal{C}} [f^*(s,r) = o^*]\right].
\end{equation}
}

\noindent{\bf (a) Multi-hop Accuracy (M-Acc).} M-Acc is used to 
compute the accuracy of the language models on multi-hop questions.
For M-Acc, we use the same settings as ~\citet{zhong2023mquake}. 
Formally, given a data instance 
$d = (\mathcal{E}, q, o, o^{*}, \mathcal{P}, \mathcal{P}^{*})$,
the calculation formula for M-Acc for the base model $f(\cdot)$ is as follows:

\begin{equation}
\mathbbm{1}\left[\bigvee_{q \in \mathcal{Q}} [f(q) = o]\right].
\end{equation}

\noindent Likewise the M-Acc for the edited model $f^{*}(\cdot)$ is 
computed as:

\begin{equation}
\mathbbm{1}\left[\bigvee_{q \in \mathcal{Q}} [f^*(q) = o^*]\right].
\end{equation}

\noindent{\bf (b) Hop-wise Accuracy (H-Acc).} H-Acc is used to 
compute the correctness of the intermediate reasoning paths for~\MQA.
In order to compute H-Acc, we follow the same settings as outlined 
by~\citet{gu2023pokemqa}. 
Given edited knowledge path $\mathcal{P}^*$, we define H-Acc as:

\begin{equation}
\mathbbm{1}\left[\bigwedge_{(s, r, o^*) \in \mathcal{P}^*}[f^*(s,r)=o^*]\right].
\end{equation}

\eat{
\noindent{\bf (d) Hop-wise Accuracy (H-Acc)}, which is used to check the correctness 
of the intermediate reasoning path for MQA. For H-Acc, we follow the same 
settings proposed by ~\citet{gu2023pokemqa}. Given edited chain of facts 
$\mathcal{C}^*$, H-Acc is defined as
\begin{equation}
\mathbbm{1}\left[\bigwedge_{(s, r, o^*) \in \mathcal{C}^*}[f^*(s,r)=o^*]\right].
\end{equation}
}

\section{Additional Experimental Results.}
\label{Appendix:results}

\subsection{Number of Hops}
\label{Appendix:res_hop}

We also compute the M-Acc for~\OurMODEL{} under varying numbers of hops.
For this, we report the performance of~\OurMODEL{} 
for~\textsc{MQuAKE-T-ar}. 
Corresponding results in Table~\ref{tab:res4_mqake_ar_hops}
compared against the baseline models show that the baseline models
experience a rapid decline in performance especially as the 
number of hops are greater than or equal to four $(\geq 4)$.
On the contrary~\OurMODEL{} yields relatively stable model
performance with the increase in the number of hops.

\begin{table}[h]
\centering
\resizebox{0.80\columnwidth}{!}{%
\begin{tabular}{lccc}
\hline
\# Hops=                    & 2-hop & 3-hop & 4-hop     \\
\hline
MeLLo                       & 85.67 & 75.67  & 23.45    \\
\OurMODEL{}                 & 92.17 & 82.25  & 64.14    \\
\hline
\end{tabular}%
}
\caption{M-Acc results for~\OurMODEL{} vs best performing baseline 
model using~\textsc{MQuAKE-T-ar} and \gptthree for 1-edited cases 
under varying number of hops.}
\label{tab:res4_mqake_ar_hops}
\end{table}

\eat{
\subsection{Edit wise Accuracy}
\label{Appendix:res_edit}
We also compute the edit-wise accuracy of~\OurMODEL{}.
Corresponding results in Table~\ref{tab:res5_mqake_ar_edits}
compared against the baseline models show that...!
\begin{table}[h]
\centering
\resizebox{0.70\columnwidth}{!}{%
\begin{tabular}{lccccc}
\hline
\# Edits =                   & 1 & 2 & 3 & 4 & All \\
\hline
FT                           & x & x & x & x & x   \\
MEND                         & x & x & x & x & x   \\
ROME                         & x & x & x & x & x   \\
MEMIT                        & x & x & x & x & x   \\
MeLLo                        & x & x & x & x & x   \\
\OurMODEL{}                  & x & x & x & x & x   \\
\hline
\end{tabular}%
}
\caption{Edit wise accuracy}
\label{tab:res5_mqake_ar_edits}
\end{table}
}

\subsection{Performance for English Language}
\label{Appendix:res_eng}
We also compared the end-performance of~\OurMODEL{} for 
English language.
Corresponding results of~\OurMODEL{} and~\textsc{MQuAKE-T}
data set compared against MeLLo~\cite{zhong2023mquake} are 
shown in Table~\ref{tab:result3_mqake_en}.
These results show~\OurMODEL{} outperforms MeLLo across 
both metrics (M-Acc, H-Acc) by a significant margin.

\begin{table}[h]
\centering
\resizebox{1.02\columnwidth}{!}{
\begin{tabular}{cccccccc}
\toprule[1.0pt]
\multirow{3}{*}{\textbf{Method}} & \multicolumn{6}{c}{\textbf{\textsc{MQuAKE-T}}}  
\\ \cmidrule{2-7} 
& \multicolumn{2}{c}{1-edited}    & \multicolumn{2}{c}{100-edited}     & \multicolumn{2}{c}{1868-edited}         
\\ \cmidrule{2-7} 
               & M-Acc  & H-Acc & M-Acc  & H-Acc   & M-Acc  & H-Acc   \\ 
\hline
\multicolumn{7}{c}{\scshape \gptthree} \\ 
\hline
MeLLo          & 77.58 & 71.13  & 82.10  & 74.51 & 73.77  & 55.41 \\
\OurMODEL{}    & \textbf{90.13} & \textbf{82.85}  & \textbf{87.02}  & \textbf{81.15} & \textbf{79.73}  & \textbf{71.78} \\

\bottomrule[1.0pt]
\end{tabular} }
\caption{Performance comparison of~\OurMODEL{} compared against MeLLo~\cite{zhong2023mquake} 
for English language. For these results, we consider a batch of $k$ instances, \textit{i.e.,} 
$k \in \{1, 100, 1868\}$ for \textsc{MQuAKE-T}. We boldface the best scores.}
\vspace{-3.7ex}
\label{tab:result3_mqake_en}
\end{table}

\newpage

\onecolumn 
\section{Prompts}
\label{Appendix:Prompts}
\subsection{Prompts for Task Decomposition ($\text{T}_\text{relation}$)}
\label{Appendix:ex2}
\begin{figure}[ht]
    \centering
    \vspace{-2.7ex}
    \includegraphics[width=0.78\columnwidth]{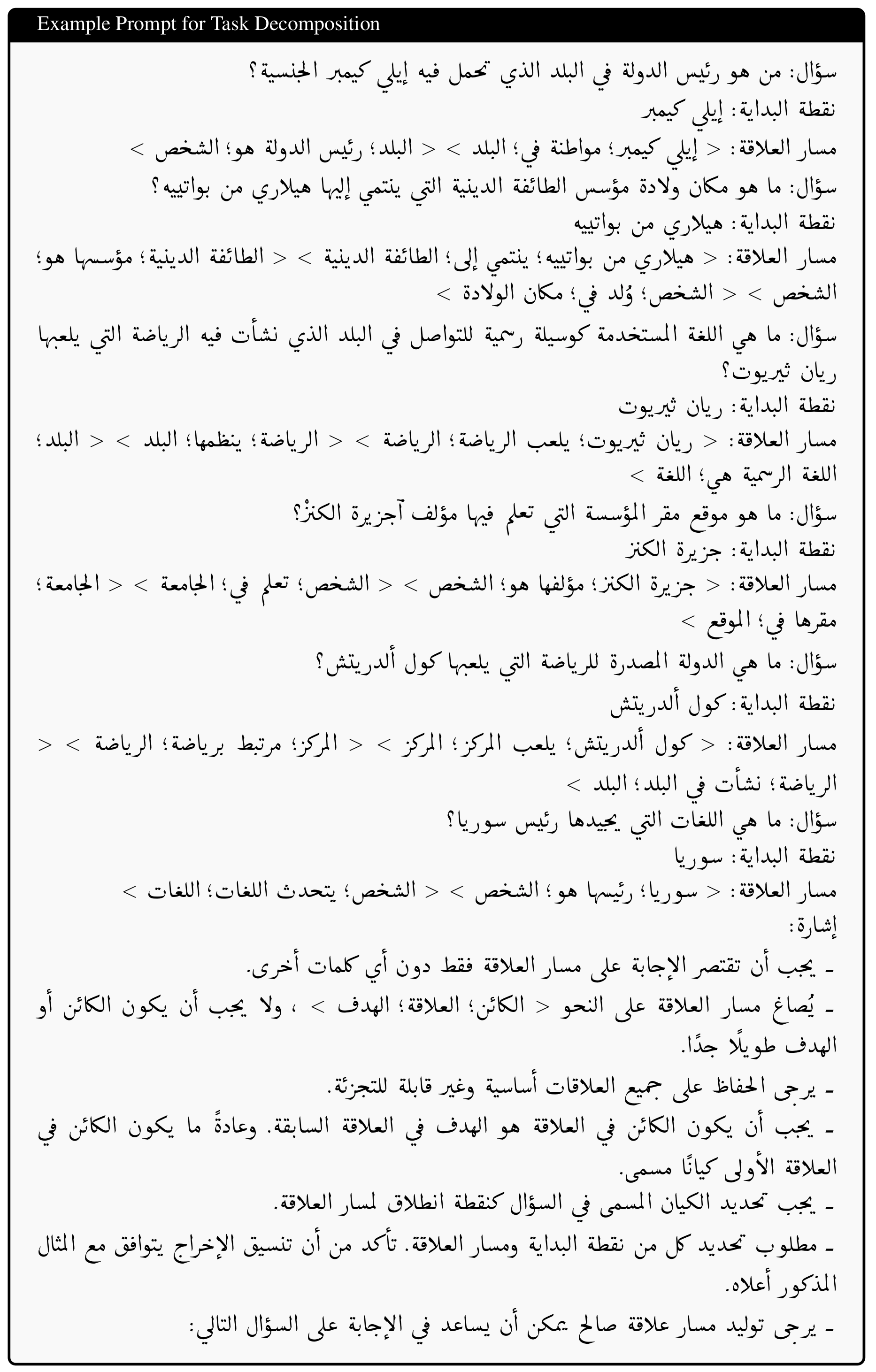}
\end{figure}

\clearpage 
\subsection{Prompts for Querying Target Model $\text{T}_\text{query}$}
\label{Appendix:ex3}

\begin{figure}[h]
    \centering
    \vspace{-2.7ex}
    \includegraphics[width=1.0\columnwidth]{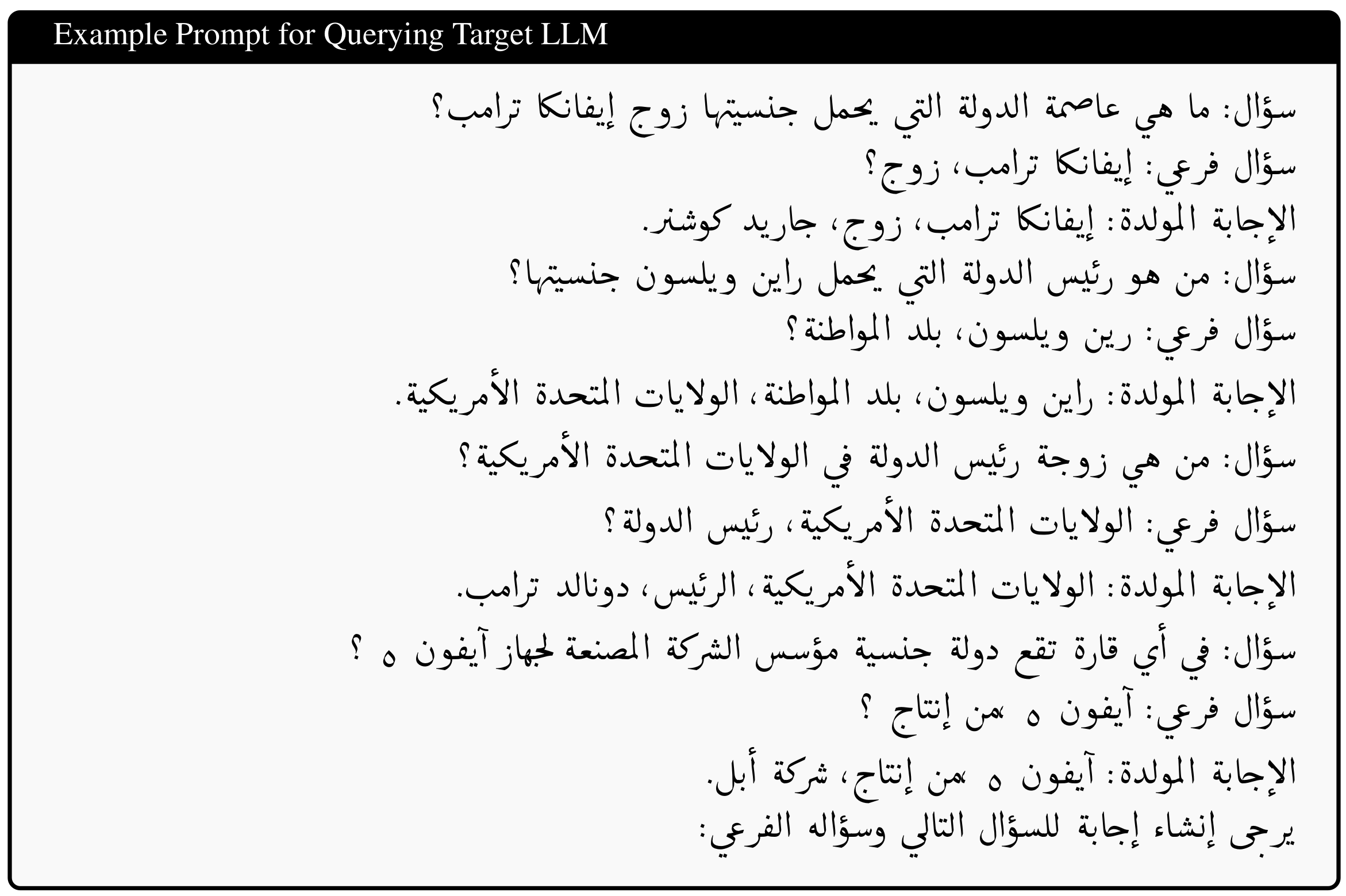}
\end{figure}


\eat{
\subsection{Single-hop QA}
\paragraph{(i) \OurDATA{}}

\paragraph{(ii) Counterfact.}
This data is a 
Counterfact dataset, used in Eva-KELLM~\citep{wu2023evakellm}

\paragraph{(iii) Know-Edit.} This data set was constructed by 
\citet{wang2023easyedit}. It is a blend of ZsRE, WikiBio, 
WikiDatacounterfact, Convsent and Sanitation data sets.

\paragraph{(iiii)~\textsc{AToKe}.} This datasets was the first temporal editing datasets constructed by ~\citet{Yin2023HistoryMT}. It consists of \textsc{AToKe-SE}, \textsc{AToKe-ME}, and \textsc{AToKe-EE}. These three respectively represent: single temporal edit (SE), multiple consecutive temporal edit (ME), extend edit (EE).

\subsection{Multi-hop QA}
}

\eat{\paragraph{(ii) \textsc{AToKe-ar}.}
This data set is..
\warn{Add details..!}}

\end{document}